\documentclass[journal]{IEEEtran}
\usepackage{amsmath,amsfonts,amssymb}
\usepackage{array}
\usepackage{booktabs}
\usepackage{multirow}
\usepackage{graphicx}
\usepackage{capt-of}
\usepackage{cite}
\usepackage{url}
\usepackage{xcolor}
\newcommand{\citep}[1]{\cite{#1}}
\newcommand{\method}{AnomalyAgent}

\usepackage{tcolorbox}
\usepackage{listings}
\usepackage{array}
\usepackage{subcaption}
\usepackage[table]{xcolor}

\newcolumntype{L}[1]{>{\raggedright\arraybackslash}p{#1}}
\newcolumntype{C}[1]{>{\centering\arraybackslash}p{#1}}

\begin{document}

\title{Agentic Ensemble with Vision-Language Models for Visual Anomaly Detection}
\author{Shangquan Sun, Wenjun Luo, Xu He, Yoonsoo Lee, and Shijian Lu
\thanks{Shangquan Sun, Wenjun Luo, Xu He, and Shijian Lu are with the College of Computing and Data Science, Nanyang Technological University, Singapore. Shijian Lu is the corresponding author (e-mail: Shijian.Lu@ntu.edu.sg).}
\thanks{Yoonsoo Lee is with Manufacturing Engineering, Hyundai Motor Company.}
\markboth{IEEE Transactions on Multimedia, 2026}{Sun \MakeLowercase{\textit{et al.}}: AnomalyAgent}
}
\maketitle

\begin{abstract}
Visual anomaly detection aims to detect abnormal objects, while anomaly samples are often scarce during model training.
Assembling multiple pretrained detectors can help by leveraging their transferable image scores and anomaly maps.
However, most existing ensemble methods apply fixed fusion rules over saved predictions which struggle when pretrained detectors clearly disagree due to different training objectives, priors, and visual representations.
To this end, we design \textbf{AnomalyAgent}, a post-training fusion framework for accurate visual anomaly detection.
AnomalyAgent aligns heterogeneous scores and maps, summarizes score- and region-level disagreement, and treats fusion as a finite action-selection problem.
It implements two types of actions: 1) score actions that determine how to fuse image-level detector confidence via consensus, robust aggregation, detector weighting, or score--map consistency; 2) map actions that fuse dense anomaly responses via consensus, detector weighting, geometric aggregation, or disagreement correction. 
A constrained vision-language model (VLM) is introduced to inspect high-conflict samples and select valid score action and map action to produce the final prediction.
We evaluate AnomalyAgent on MVTec AD and VisA as primary benchmarks, together with seven additional public anomaly benchmarks and Real-IAD for large-scale score-level evaluation.
Across these benchmarks, AnomalyAgent consistently improves over equal-weight fusion and achieves the best mean image AUROC and pixel AP among the compared fusion methods on MVTec AD and VisA, while demonstrating robust transfer across additional anomaly domains.
\end{abstract}
\begin{IEEEkeywords}
Visual anomaly detection, post-training ensemble, vision-language model, anomaly localization.
\end{IEEEkeywords}

\section{Introduction}
\label{sec:intro}

Visual anomaly detection (VAD) aims to detect anomalous samples and localize abnormal regions under limited anomaly supervision, where anomalies are often rare, diverse, and difficult to characterize exhaustively.
The prevalent approach mitigates the lack of anomaly supervision by leveraging pretrained anomaly detectors that provide transferable image-level anomaly scores and pixel-level localization maps.
It substantially improves VAD performance by exploiting vision-language representations~\cite{jeong2023winclip,zhou2023anomalyclip,qu2024vcp}, feature reconstruction~\cite{guo2025dinomaly}, memory modeling~\cite{roth2022towards,defard2021padim}, state-space architectures~\cite{he2024mambaad,zhang2025wave}, foundation-model priors~\cite{oquab2023dinov2}, etc.
However, different detectors, pretrained under different objectives and priors, often exhibit different failure modes: some may be sensitive to harmless texture variations, some may miss small structural anomalies, and some may produce diffused localization~\cite{liu2024survey,cui2023survey}.
Their very diverse predictions suggest that pretrained detectors learn complementary evidence and that properly assembling them could outperform any individual detector.

Different ensemble approaches have been explored, including classical strategies such as score averaging~\cite{dietterich2000ensemble}, weighted voting~\cite{aggarwal2016outlier,rayana2016less,boateng2023unsupervised}, and stacking~\cite{wolpert1992stacked}.
These methods can improve the detection robustness when validation labels or task-specific fitting are available.
However, they struggle while handling label-free post-training visual inspection, where anomaly labels are unavailable, base detectors are fixed, and the fusion module must handle both image-level scores and dense anomaly maps without retraining.
Under such a scenario, the most accessible post-training baseline is equal-weight averaging that normalizes and averages the scores and anomaly maps of all assembled detectors.
Despite being simple and deterministic, this strategy assumes that every detector is equally reliable for every image and every spatial region.
As a result, an isolated false response from one detector will be aggregated into the fused anomaly map and distort it, and a subtle but valid anomaly detected by only one expert can be diluted by the others' predictions.
The key challenge thus becomes evidence arbitration: an ensemble method should understand which detector to trust, which response to reject, and how to resolve inter-detector prediction conflicts.

To address this problem, we propose \textbf{AnomalyAgent}, a structured post-training fusion framework for visual anomaly detection.
Rather than modifying the pretrained detectors or training an additional meta-detector, AnomalyAgent handles VAD after the pretrained detectors have produced their prediction scores and dense anomaly maps.
It normalizes heterogeneous outputs, builds a disagreement-aware evidence state, and invokes a constrained vision-language model (VLM) when the conflict score exceeds a predefined threshold.
Here the VLM acts as a single-step policy selector over finite fusion actions, instead of a free-form mask generator or a multi-turn tool-using agent---every valid decision remains grounded in the outputs of the pretrained detectors.
For image-level prediction, the available score actions include consensus, robust aggregation, detector weighting, and score--map consistency, allowing the VLM to determine how detector confidence should be combined.
For pixel-level localization, the map actions include consensus, detector weighting, geometric aggregation, and disagreement correction, enabling the VLM to suppress unreliable responses or preserve complementary spatial evidence.
The selected score and map actions are then applied to the fixed detector outputs to produce the final anomaly score and localization map.

Our main contributions are summarized as follows:
\begin{itemize}
    \item We formulate label-free transductive post-training anomaly-detector fusion as conflict-conditioned selection over deterministic score- and map-level operators, without updating the base detectors.

    \item We introduce a constrained VLM policy that makes each intervention traceable and prevents free-form anomaly generation through selection over valid detector-grounded fusion actions.

    \item Across all six MVTec AD and VisA benchmark--pool configurations, AnomalyAgent consistently improves over equal-weight fusion and achieves the best mean image AUROC and pixel AP among the compared fusion methods on both primary benchmarks, together with robust transfer across seven additional public benchmarks and effective large-scale evaluation on Real-IAD.
\end{itemize}
\section{Related Work}
\label{sec:related}

\paragraph{Visual anomaly detection.}
Visual anomaly detection aims to identify anomalous images and localize abnormal regions under limited anomaly supervision.
MVTec AD~\citep{bergmann2019mvtec} and VisA~\citep{zou2022spot} are widely used benchmarks, while SDD/KSDD~\citep{tabernik2020segmentation}, Real-IAD~\citep{wang2024real}, MVTec 3D-AD~\citep{bergmann2021mvtec}, Kvasir-SEG~\citep{jha2019kvasir}, and endoluminal scene segmentation benchmarks~\citep{vazquez2017benchmark} extend evaluation across surface, 3D, medical, and other anomaly settings.

Existing methods are commonly grouped into embedding-, discrimination-, and reconstruction-based approaches~\citep{liu2024survey,cui2023survey}.
Embedding-based methods model normal feature distributions or memory banks with pretrained representations, including PaDiM~\citep{defard2021padim}, PatchCore~\citep{roth2022towards}, CS-Flow~\citep{rudolph2022fully}, PyramidFlow~\citep{lei2023pyramidflow}, and VQ-Flow~\citep{zhou2025vq}.
Discrimination-based methods learn decision boundaries from synthetic or predefined pseudo anomalies, including CutPaste~\citep{li2021cutpaste}, DRAEM~\citep{zavrtanik2021draem}, SimpleNet~\citep{liu2023simplenet}, open-set supervised anomaly detection~\citep{ding2022catching}, and recent open-set visual anomaly recognition methods~\citep{yang2026open}.
Reconstruction- and distillation-based methods rely on reconstruction errors, teacher--student discrepancies, feature reconstruction, or inversion, including autoencoder-based models~\citep{bengio2013generalized,vincent2008extracting}, Uninformed Students~\citep{bergmann2020uninformed}, Reverse Distillation~\citep{deng2022anomaly}, UniAD~\citep{you2022unified}, FastRecon~\citep{fang2023fastrecon}, OneNIP~\citep{gao2024learning}, InvAD~\citep{zhang2024invad}, Dinomaly~\citep{guo2025dinomaly}, self-supervised masking~\citep{huang2022self}, PADNet~\citep{yang2025padnet}, and memory-based coarse-to-fine feature fusion~\citep{zhu2026anomaly}.
Recent methods further explore state-space architectures, multimodal sensing, and foundation-model priors for stronger multi-class or cross-domain generalization~\citep{he2024mambaad,zhang2025wave,oquab2023dinov2,liu2025multimodal}.
Different from these works, we do not design another base detector, but combine image-level scores and dense anomaly maps from fixed pretrained detectors at test time.

\paragraph{Vision-language anomaly detectors.}
Large vision-language models such as CLIP~\citep{radford2021learning} have improved anomaly detection transfer across categories and domains.
WinCLIP~\citep{jeong2023winclip} uses normal/abnormal text prompts and multi-scale window features, while AnomalyCLIP~\citep{zhou2023anomalyclip}, AdaCLIP~\citep{cao2024adaclip}, and VCP-CLIP~\citep{qu2024vcp} adapt anomaly, textual, or visual-context prompts for classification and segmentation.
Other studies exploit language guidance and vision-language feature interaction~\citep{chen2024improving,zhang2026tunclip}.
Beyond prompt adaptation, ACR~\citep{li2023zero} and MuSc~\citep{li2024musc} use distributional or mutual-scoring mechanisms, while PromptAD~\citep{li2024promptad}, one-normal-prompt learning~\citep{gao2024learning}, and InCtrl~\citep{zhu2024toward} study prompt-based or in-context adaptation with limited normal samples.
These methods mainly improve a single detector or adapt a vision-language backbone.
In contrast, \method{} uses a VLM as a constrained selector over multiple detector outputs, choosing only actions grounded in existing anomaly evidence rather than generating anomaly maps freely.

\paragraph{Post-training ensemble fusion.}
Ensemble learning has been widely studied through averaging, weighted voting, selective ensembles, and stacking~\citep{dietterich2000ensemble,wolpert1992stacked,aggarwal2016outlier,rayana2016less,boateng2023unsupervised,sun2024ensir}.
Recent anomaly-detection systems also explore feature- and modality-level fusion, including memory-guided multimodal anomaly detection and coarse-to-fine feature fusion~\citep{liu2025multimodal,zhu2026anomaly}.
These approaches combine complementary evidence, but generally fuse representations within jointly designed detectors.
Post-training fusion instead operates on fixed detectors, often without anomaly labels, and must combine both scalar image scores and dense anomaly maps despite image- and region-dependent detector reliability.
Equal-weight averaging is a useful baseline, but assumes uniform reliability across images, regions, and output types.
\method{} instead conditions fusion on detector disagreement and structured visual evidence while retaining averaging as a safe reference.

\paragraph{Vision-language agents for expert arbitration.}
Recent language and vision-language agents reason over observations, intermediate states, and external tools~\citep{yao2022react,wu2023visual,suris2023vipergpt,yang2023mm,hu2025survey,liu2026last}.
Related anomaly-detection work has explored multimodal evidential reasoning and training-free VLM decision processes, including open-world video anomaly detection and uncertainty-guided hierarchical retrieval~\citep{huang2025multimodal,zheng2026training}.
Direct VLM mask generation, however, can be unstable or weakly grounded for quantitative dense localization.
\method{} therefore restricts the VLM to selecting score- and map-level actions for high-conflict cases from the image, detector scores and maps, candidate regions, disagreement summaries, and a bounded action set.
It cannot directly draw a mask or emit a new anomaly score, so the final prediction remains a reproducible transformation of detector outputs.
\section{Method}
\label{sec:method}

\begin{figure*}[t]
    \centering
    \includegraphics[width=1\linewidth]{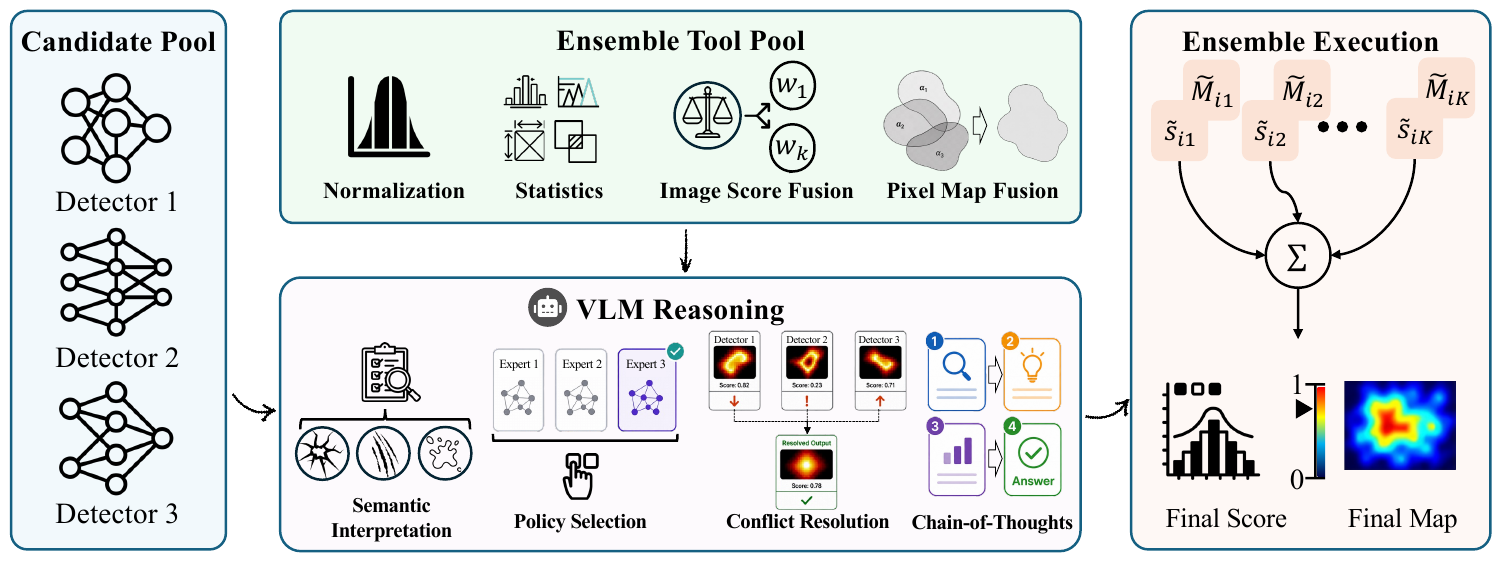}
    \caption{\method{} overview. Pretrained anomaly detectors produce image-level scores and dense anomaly maps. Their outputs are normalized and summarized into a structured evidence state. 
    The VLM is invoked when the conflict score exceeds a fixed threshold and chooses among fixed image- and pixel-level fusion actions.}
    \label{fig:architecture}
    \vspace{-2mm}
\end{figure*}
\subsection{Post-Training Anomaly-Detector Ensemble}

We consider a post-training ensemble of fixed anomaly detectors for visual anomaly detection.
For a test image $x_i$, detector $k\in\{1,\ldots,K\}$ outputs an image-level anomaly score $s_{ik}\in\mathbb{R}$ and a dense anomaly map $M_{ik}\in\mathbb{R}^{H_i\times W_i}$, where $H_i$ and $W_i$ are the map height and width.
The detector pool is fixed: \method{} does not update detector weights, prompts, memory banks, or reconstruction modules.
The fusion problem is to map the prediction set
\begin{equation}
    \mathcal{P}_i=\{(s_{ik},M_{ik})\}_{k=1}^{K}
\end{equation}
to a final image-level anomaly score $\hat{s}_i$ and pixel-level anomaly map $\hat{M}_i$.

Because pretrained detectors may have different score ranges and map calibrations, we first apply group-wise robust normalization:
\begin{equation}
    \tilde{s}_{ik}=\mathrm{Norm}_{g_i}(s_{ik}),\qquad
    \tilde{M}_{ik}=\mathrm{Norm}_{g_i}(M_{ik}),
\end{equation}
where $g_i$ denotes the evaluation group of image $x_i$, such as object category, dataset category, or benchmark-defined subset when such metadata is available.
$\mathrm{Norm}_{g_i}$ denotes percentile-clipped min--max scaling within that group.
This label-free transductive calibration uses detector outputs and metadata only, and does not require anomaly labels.
The standard averaging-based ensemble is
\begin{equation}
    \bar{s}_i=\frac{1}{K}\sum_{k=1}^{K}\tilde{s}_{ik},\qquad
    \bar{M}_i=\frac{1}{K}\sum_{k=1}^{K}\tilde{M}_{ik}.
\end{equation}
Although $\bar{s}_i$ and $\bar{M}_i$ provide simple and deterministic references, they assume that all detectors are equally reliable for every image, region, and output type.
In contrast, \method{} replaces this fixed assumption with image-conditioned evidence arbitration.

\subsection{Structured Evidence State}

For each image, \method{} summarizes both the detector predictions and their internal disagreement into a structured evidence representation.
We first characterize pixel-level disagreement using the detector response range:
\begin{equation}
    R_i(u)=\max_{k}\tilde{M}_{ik}(u)-\min_{k}\tilde{M}_{ik}(u),
\end{equation}
where $u$ denotes a pixel location. A large $R_i(u)$ indicates that the detectors produce substantially different responses at that location. However, raw disagreement may be dominated by isolated noisy pixels. To emphasize compact high-response regions, we further define the top-response functional
\begin{equation}
    T_q(M)=\frac{1}{|\Omega_q(M)|}
    \sum_{u\in\Omega_q(M)}M(u),
\end{equation}
where $\Omega_q(M)$ denotes the set containing the top-$q$ fraction of pixels ranked by their responses in map $M$. Compared with the maximum response, $T_q$ is less sensitive to isolated spikes, while remaining more responsive to compact abnormal regions than a global average.
Based on these quantities, we construct a label-free collection of derived evidence statistics
\begin{equation}
    \Delta_i =
    \left(
    R_i,
    T_q(R_i),
    \{T_q(\tilde{M}_{ik})\}_{k=1}^{K},
    T_q(\bar{M}_i),
    \mathcal{C}_i
    \right),
\end{equation}
where $\mathcal{C}_i$ denotes the candidate high-response regions extracted from the fused and detector-specific anomaly maps. Thus, $\Delta_i$ summarizes spatial disagreement, detector-specific high-response statistics, fused-map responses, and candidate regions without using anomaly labels.
The complete structured evidence state is then defined as
\begin{equation}
    E_i=
    \left(
    x_i,
    \tilde{s}_{i1:K},
    \tilde{M}_{i1:K},
    \bar{s}_i,
    \bar{M}_i,
    \Delta_i
    \right).
\end{equation}
Therefore, $E_i$ contains the original image, normalized detector scores and maps, the equal-weight fusion reference, and derived disagreement evidence. This representation provides the information used by the subsequent conflict trigger and constrained VLM fusion policy.

\begin{table*}[t]
\centering
\footnotesize
\renewcommand{\arraystretch}{1.15}
\setlength{\tabcolsep}{18pt}
\caption{Detector-grounded fusion-action families for the constrained VLM policy in image-score and pixel-map arbitration.}
\label{tab:action-space}
\begin{tabular}{lll}
\toprule
Family & Image-score operator $F_s$ & Pixel-map operator $F_m$\\
\midrule
Consensus & $\frac{1}{K}\sum_k\tilde{s}_{ik}$ & $\frac{1}{K}\sum_k\tilde{M}_{ik}$\\
Robust consensus & $\mathrm{median}_k(\tilde{s}_{ik})$ & $\mathrm{median}_k(\tilde{M}_{ik})$\\
Detector weighting & $\sum_k w_k\tilde{s}_{ik}$ & $\sum_k w_k\tilde{M}_{ik}$\\
Score--map consistency & $(1-\alpha)\sum_k w_k\tilde{s}_{ik}+\alpha T_q(M)$ & --\\
Disagreement correction & -- & $\mathrm{clip}(M-\eta R_i,0,1)$ or $\mathrm{clip}(M+\eta R_i,0,1)$\\
Geometric consensus & -- & $\exp\!\left(\frac{1}{K}\sum_k\log(\tilde{M}_{ik}+\epsilon)\right)$\\
\bottomrule
\end{tabular}
\end{table*}

\subsection{Thresholded High-Conflict Agent Invocation}

To improve efficiency, \method{} invokes the VLM when the pretrained detectors exhibit substantial internal disagreement. Before VLM call, we first compute a deterministic preliminary fusion
\begin{equation}
    (s_i^0,M_i^0)=
    \left(
    F_s(a^{s,0};\tilde{s}_{i1:K},\tilde{M}_{i1:K}),
    F_m(a^{m,0};\tilde{M}_{i1:K})
    \right),
\end{equation}
where $F_s$ and $F_m$ denote the image-score and pixel-map fusion operators, respectively, and $a^{s,0}$ and $a^{m,0}$ are predefined fallback actions selected from the same finite action space used by the VLM. The preliminary fusion is retained when the VLM is not invoked or when its output cannot be parsed, while equal-weight fusion $(\bar{s}_i,\bar{M}_i)$ serves as a neutral reference for measuring disagreement.
We then define a label-free conflict score
\begin{equation}
\label{eq:conflict}
    C_i =
    \lambda_s D_s(i)
    +\lambda_m D_m(i)
    +\lambda_p D_p(i)
    +\lambda_r D_r(i),
\end{equation}
where $\lambda_s,\lambda_m,\lambda_p,\lambda_r\geq0$ are hyperparameters controlling the relative contributions of different disagreement cues. These cues are defined as
\begin{align}
    D_s(i)&=\mathrm{Std}_{k}(\tilde{s}_{ik}),&
    D_m(i)&=T_q(R_i), \nonumber\\
    D_p(i)&=\left|T_q(M_i^0)-T_q(\bar{M}_i)\right|,&
    D_r(i)&=\left|s_i^0-\bar{s}_i\right|.
\end{align}
Here, $D_s$ measures image-score disagreement across detectors, $D_m$ summarizes spatial disagreement in high-response regions, $D_p$ measures the deviation of the preliminary fused map from the equal-weight reference, and $D_r$ measures the corresponding deviation at the image-score level. These terms are not intended to estimate anomaly likelihood; instead, they quantify internal inconsistency within the detector ensemble.
After that, the VLM is invoked when the conflict score exceeds a threshold:
\begin{equation}
    \mathcal{I}_{\tau}=\{i\,:\,C_i\geq\tau\}.
\end{equation}
Samples in $\mathcal{I}_{\tau}$ are sent to the VLM for fusion-action selection, while the remaining samples keep the preliminary fusion result. 
We use a fixed intervention threshold, which does not use anomaly labels.

This conflict-triggered design separates anomaly likelihood from arbitration need: a highly anomalous image does not require additional reasoning if the detectors already agree, whereas even a low-scoring image may require arbitration when the detector outputs are inconsistent. By reserving VLM inference for high-conflict cases, \method{} avoids unnecessary computation on samples for which the detector ensemble already provides consistent evidence.
\begin{table*}[t]
\centering
\footnotesize
\setlength{\tabcolsep}{20pt}
\renewcommand{\arraystretch}{0.8}
\caption{Complete comparison on MVTec AD across the three detector pools.
All values are percentages, and the best result within each detector-pool block is boldfaced.}
\label{tab:primary-mvtec}
\begin{tabular}{lrrrrrr}
\toprule
& \multicolumn{3}{c}{Image-level} & \multicolumn{3}{c}{Pixel-level} \\
\cmidrule(lr){2-4} \cmidrule(lr){5-7}
Method & AUROC & AP & F1 & AUROC & AP & F1 \\
\midrule
\midrule

\multicolumn{7}{c}{\textit{CLIP-based detector pool}} \\
\midrule
AdaCLIP~\cite{cao2024adaclip}
& 99.17 & 99.66 & 98.34
& \textbf{98.99} & \textbf{72.31} & \textbf{71.47} \\
AnomalyCLIP~\cite{zhou2023anomalyclip}
& 91.56 & 96.36 & 92.75
& 91.05 & 31.28 & 38.37 \\
WinCLIP~\cite{jeong2023winclip}
& 93.56 & 96.86 & 93.52
& 96.40 & 49.74 & 54.21 \\
Equal-weight~\cite{dietterich2000ensemble}
& 98.80 & 99.50 & 97.95
& 97.83 & 64.43 & 64.87 \\
Inverse-distance~\cite{shepard1968two}
& 98.79 & 99.43 & 97.74
& 97.67 & 62.18 & 62.65 \\
Voting~\cite{aggarwal2016outlier,rayana2016less,boateng2023unsupervised}
& 97.13 & 98.68 & 96.30
& 94.98 & 51.06 & 57.25 \\
Stacking~\cite{wolpert1992stacked}
& 99.23 & 99.68 & 98.41
& 98.07 & 69.03 & 69.27 \\
\rowcolor{gray!10}
\method{}
& \textbf{99.25} & \textbf{99.69} & \textbf{98.45}
& \textbf{98.99} & \textbf{72.31} & \textbf{71.47} \\

\midrule
\midrule

\multicolumn{7}{c}{\textit{Feature-reconstruction detector pool}} \\
\midrule
MambaAD~\cite{he2024mambaad}
& 98.54 & 99.52 & 97.77
& 97.64 & 51.85 & 58.18 \\
OneNIP~\cite{gao2024learning}
& 97.98 & 99.35 & 97.60
& 97.85 & 58.47 & 60.75 \\
InvAD~\cite{zhang2024invad}
& 95.53 & 97.91 & 94.88
& 95.93 & 35.47 & 43.52 \\
Equal-weight~\cite{dietterich2000ensemble}
& 98.98 & 99.69 & 98.28
& 98.18 & 59.67 & 61.52 \\
Inverse-distance~\cite{shepard1968two}
& 98.93 & 99.68 & 98.45
& 98.16 & 58.74 & 61.30 \\
Voting~\cite{aggarwal2016outlier,rayana2016less,boateng2023unsupervised}
& 96.30 & 98.24 & 96.06
& 95.92 & 44.09 & 54.39 \\
Stacking~\cite{wolpert1992stacked}
& 98.99 & 99.69 & 98.31
& 98.08 & 57.83 & 60.45 \\
\rowcolor{gray!10}
\method{}
& \textbf{99.11} & \textbf{99.74} & \textbf{98.62}
& \textbf{98.27} & \textbf{60.99} & \textbf{62.66} \\

\midrule
\midrule

\multicolumn{7}{c}{\textit{Hybrid detector pool}} \\
\midrule
OneNIP~\cite{gao2024learning}
& 97.98 & 99.35 & 97.60
& 97.85 & 58.38 & 60.69 \\
InvAD~\cite{zhang2024invad}
& 95.53 & 97.91 & 94.88
& 95.93 & 35.41 & 43.47 \\
AdaCLIP~\cite{cao2024adaclip}
& 99.17 & 99.66 & 98.34
& 98.99 & 72.28 & 71.44 \\
Equal-weight~\cite{dietterich2000ensemble}
& 99.65 & 99.90 & 99.04
& 98.93 & 72.98 & 72.04 \\
Inverse-distance~\cite{shepard1968two}
& 99.62 & 99.90 & 99.09
& 98.87 & 70.70 & 69.29 \\
Voting~\cite{aggarwal2016outlier,rayana2016less,boateng2023unsupervised}
& 97.87 & 98.90 & 97.55
& 93.25 & 51.72 & 62.17 \\
Stacking~\cite{wolpert1992stacked}
& 99.71 & \textbf{99.92} & 99.21
& 98.85 & 72.63 & 72.13 \\
\rowcolor{gray!10}
\method{}
& \textbf{99.72} & \textbf{99.92} & \textbf{99.31}
& \textbf{99.13} & \textbf{74.61} & \textbf{73.45} \\

\bottomrule
\end{tabular}
\vspace{-2mm}
\end{table*}

\begin{figure*}[!t]
    \centering
    \setlength{\tabcolsep}{1.5pt}
    \renewcommand{\arraystretch}{1.0}

    \begin{tabular}{@{}*{9}{c}@{}}

        \multicolumn{9}{c}{
            \small\textbf{(a) MVTec AD, CLIP-based detector pool}
            \quad
            \scriptsize
            (Detector 1: AdaCLIP, Detector 2: AnomalyCLIP, Detector 3: WinCLIP)
        }
        \\[0.0mm]

        \includegraphics[width=0.105\textwidth]{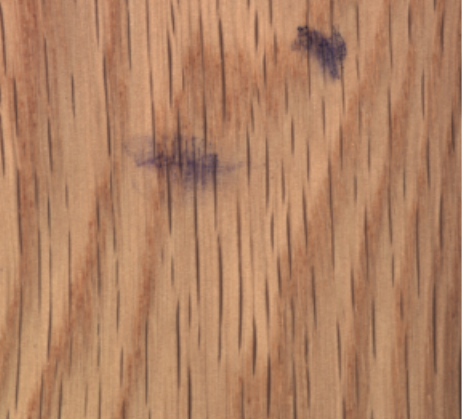}
        &
        \includegraphics[width=0.105\textwidth]{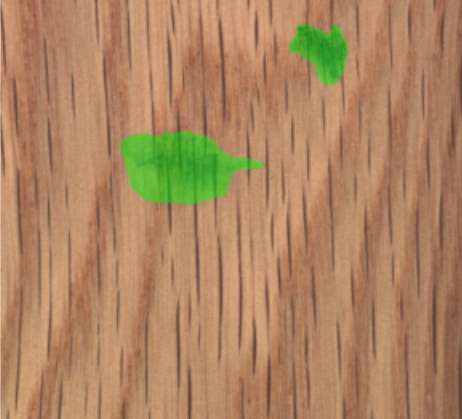}
        &
        \includegraphics[width=0.105\textwidth]{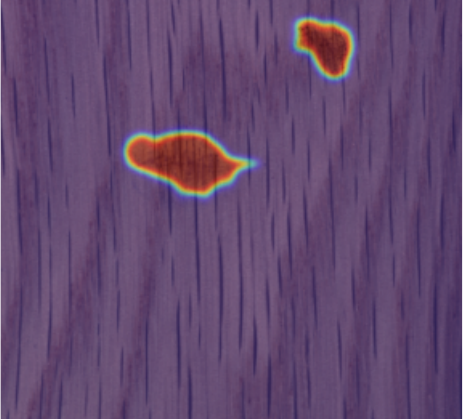}
        &
        \includegraphics[width=0.105\textwidth]{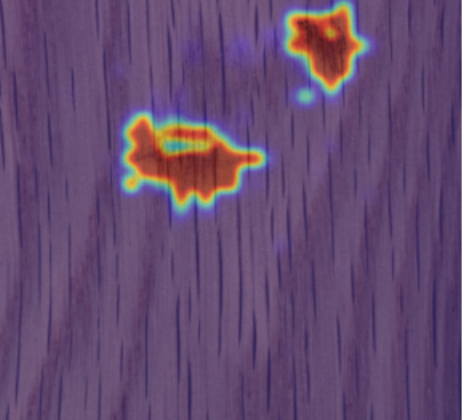}
        &
        \includegraphics[width=0.105\textwidth]{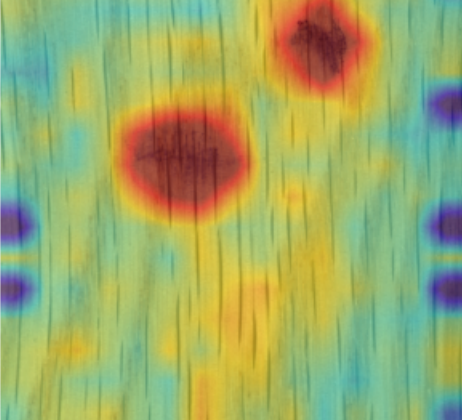}
        &
        \includegraphics[width=0.105\textwidth]{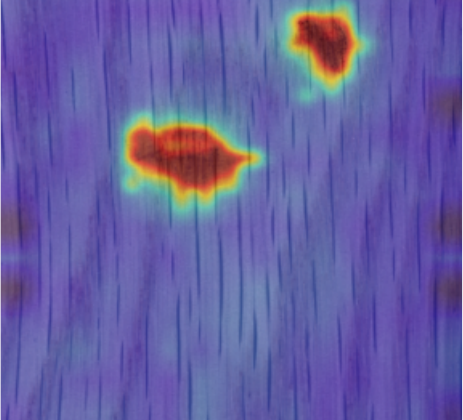}
        &
        \includegraphics[width=0.105\textwidth]{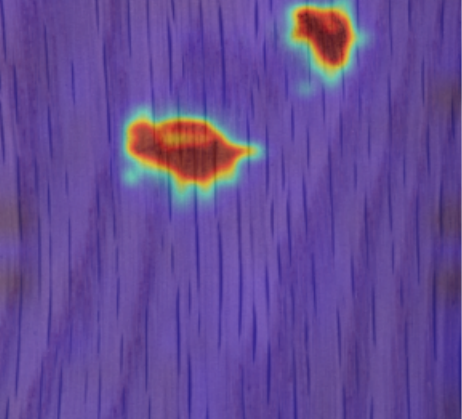}
        &
        \includegraphics[width=0.105\textwidth]{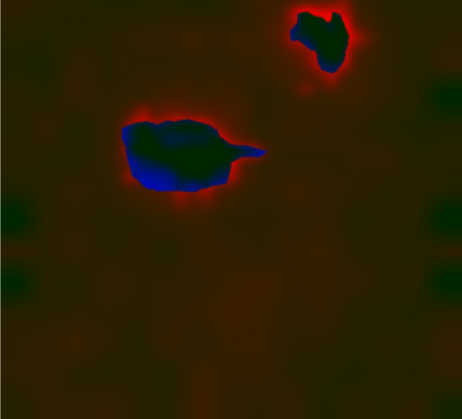}
        &
        \includegraphics[width=0.105\textwidth]{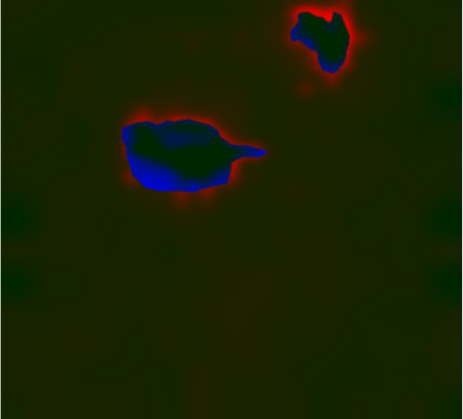}
        \\[0mm]

        \multicolumn{9}{c}{
            \small\textbf{(b) MVTec AD, hybrid detector pool}
            \quad
            \scriptsize
            (Detector 1: AdaCLIP, Detector 2: OneNIP, Detector 3: InvAD)
        }
        \\[0.mm]

        \includegraphics[width=0.105\textwidth]{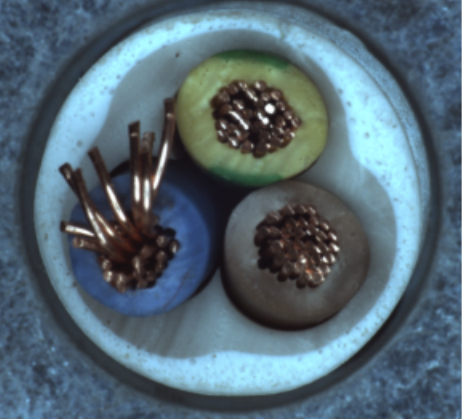}
        &
        \includegraphics[width=0.105\textwidth]{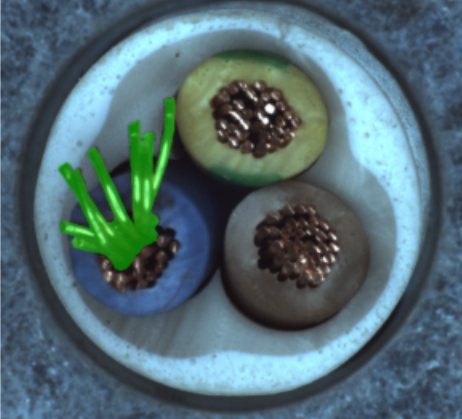}
        &
        \includegraphics[width=0.105\textwidth]{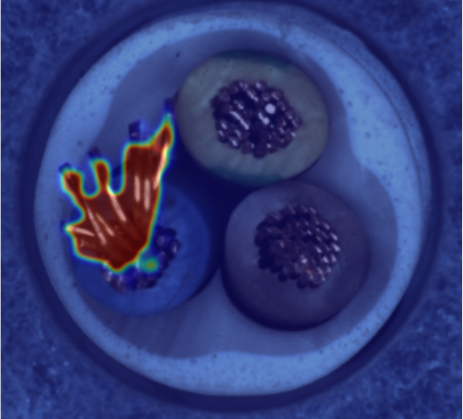}
        &
        \includegraphics[width=0.105\textwidth]{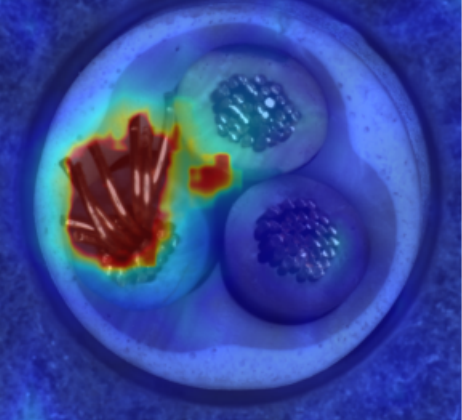}
        &
        \includegraphics[width=0.105\textwidth]{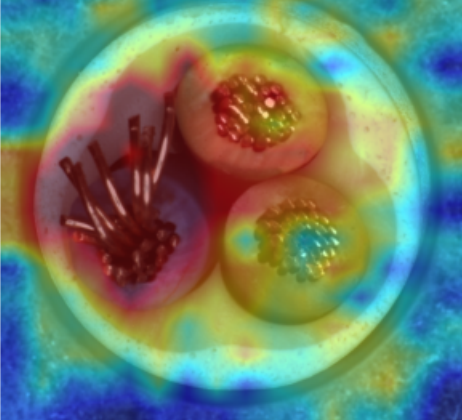}
        &
        \includegraphics[width=0.105\textwidth]{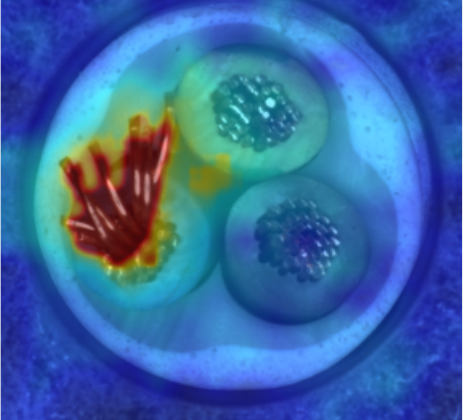}
        &
        \includegraphics[width=0.105\textwidth]{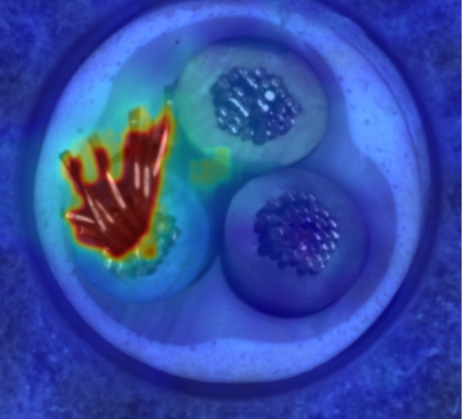}
        &
        \includegraphics[width=0.105\textwidth]{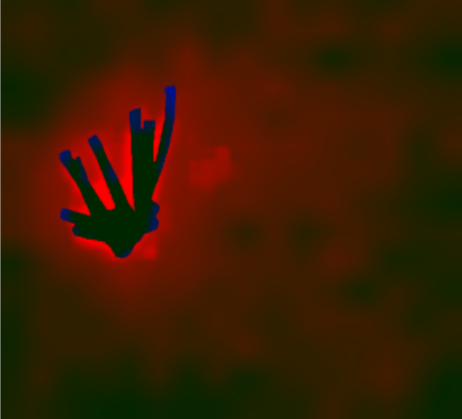}
        &
        \includegraphics[width=0.105\textwidth]{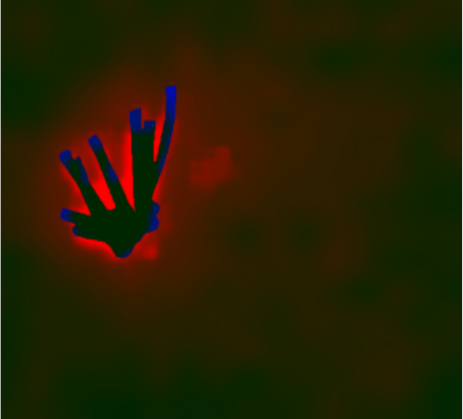}
        \\[0mm]

        \multicolumn{9}{c}{
            \small\textbf{(c) VisA, hybrid detector pool}
            \quad
            \scriptsize
            (Detector 1: AdaCLIP, Detector 2: OneNIP, Detector 3: InvAD)
        }
        \\[-2.mm]
        \includegraphics[width=0.105\textwidth]{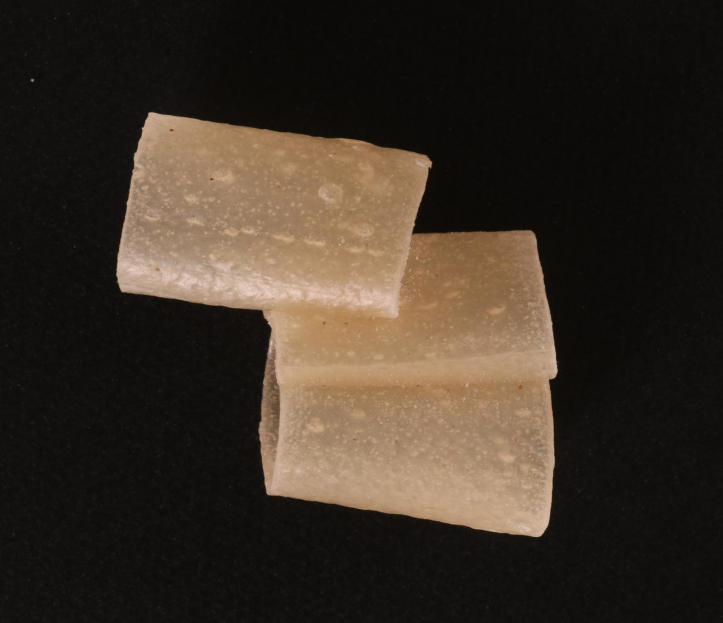}
        &
        \includegraphics[width=0.105\textwidth]{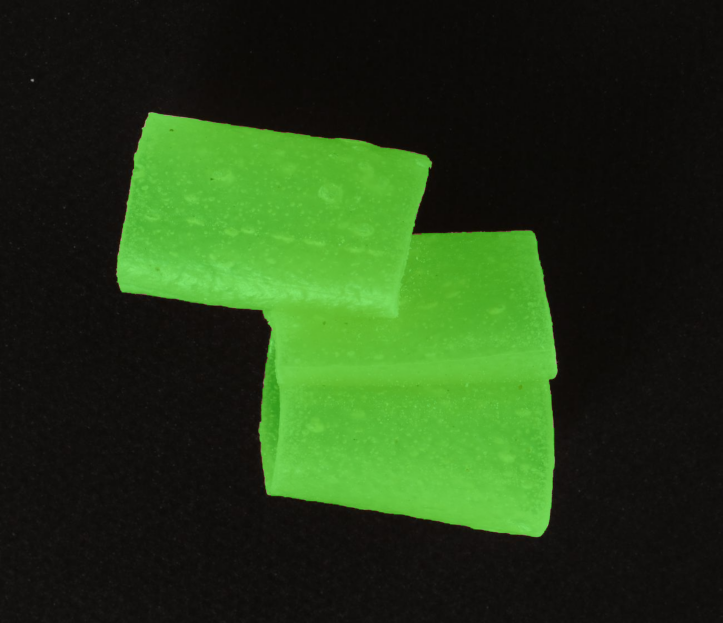}
        &
        \includegraphics[width=0.105\textwidth]{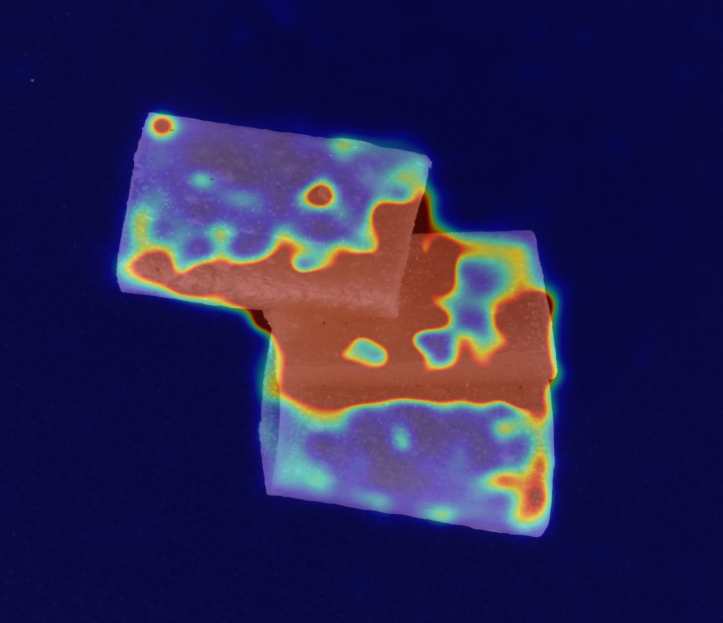}
        &
        \includegraphics[width=0.105\textwidth]{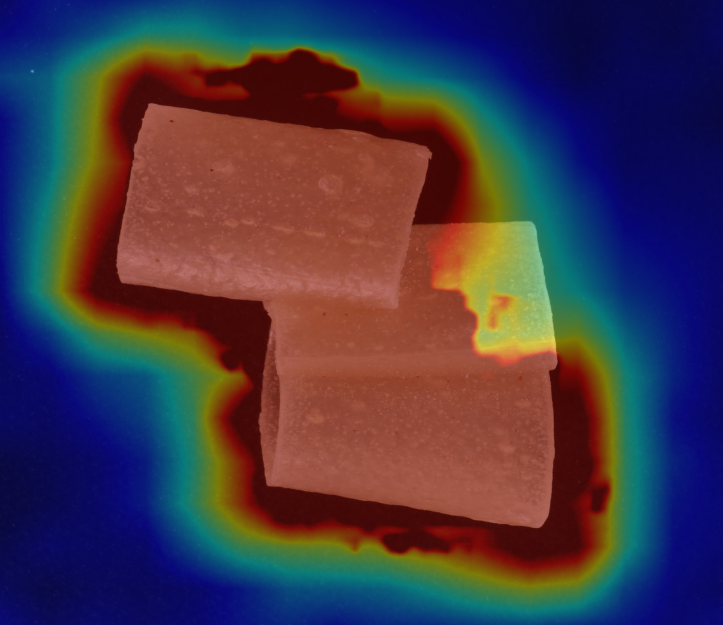}
        &
        \includegraphics[width=0.105\textwidth]{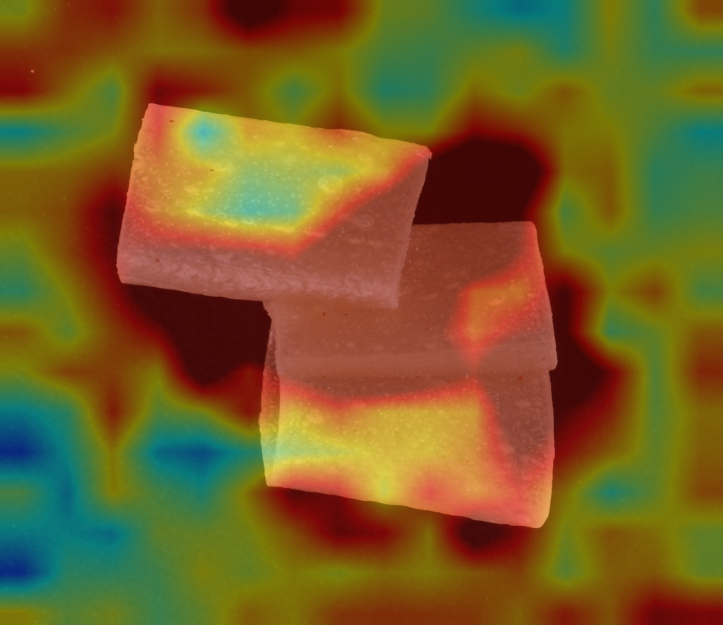}
        &
        \includegraphics[width=0.105\textwidth]{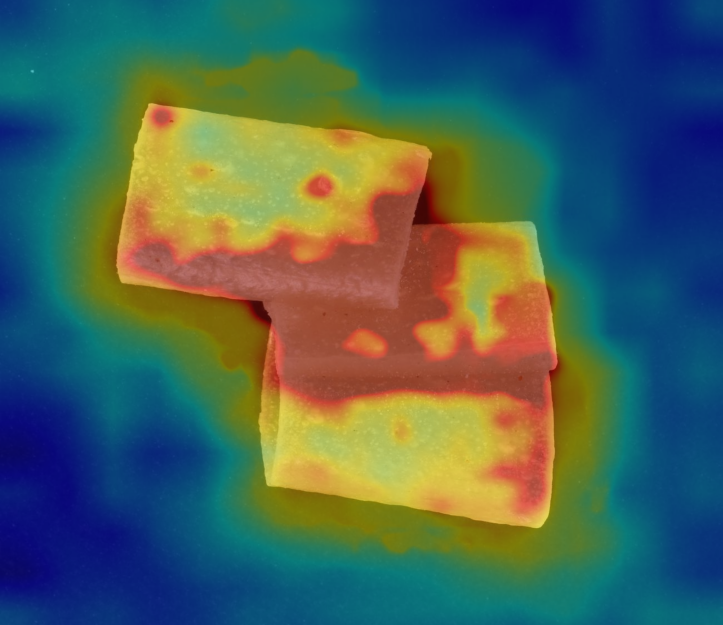}
        &
        \includegraphics[width=0.105\textwidth]{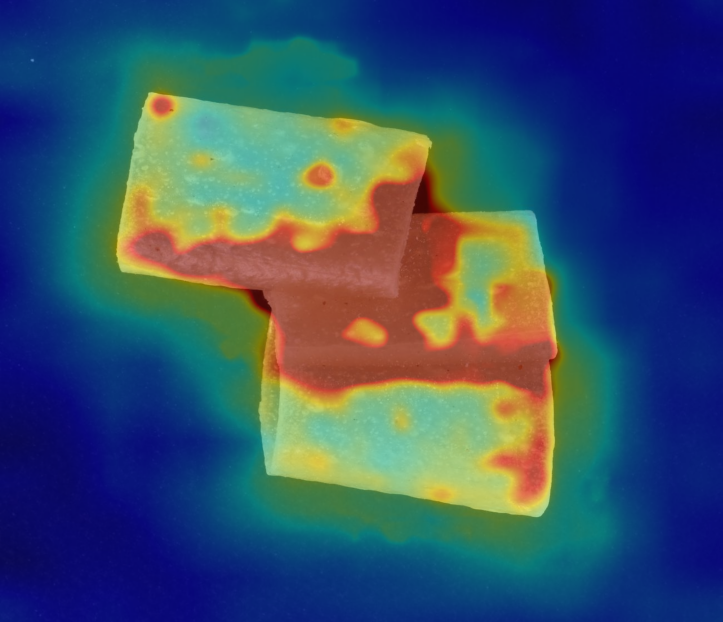}
        &
        \includegraphics[width=0.105\textwidth]{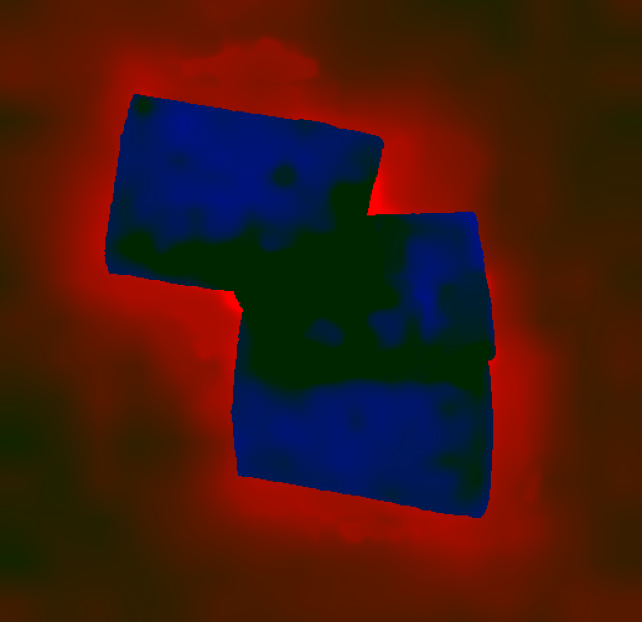}
        &
        \includegraphics[width=0.105\textwidth]{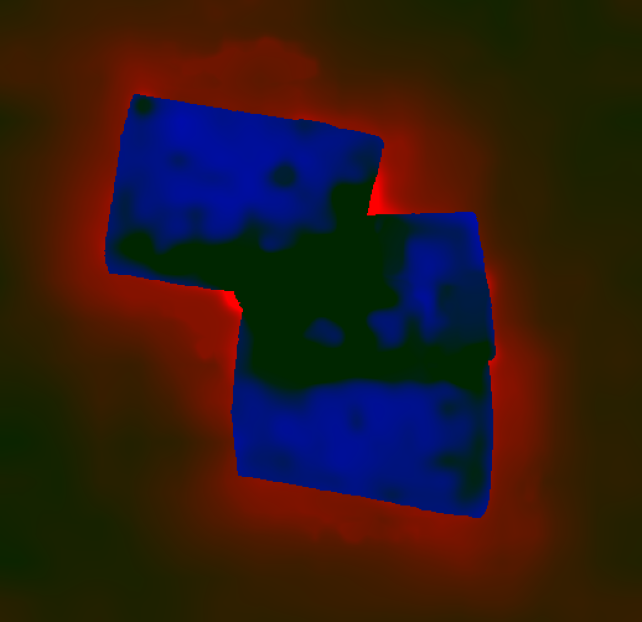}
        \\[0mm]

        \scriptsize\textbf{Input}
        &
        \scriptsize\textbf{Ground Truth}
        &
        \scriptsize\textbf{Detector 1}
        &
        \scriptsize\textbf{Detector 2}
        &
        \scriptsize\textbf{Detector 3}
        &
        \scriptsize\textbf{Equal-weight}
        &
        \scriptsize\textbf{Ours}
        &
        \scriptsize\textbf{Error (Equal-weight)}
        &
        \scriptsize\textbf{Error (Ours)}

    \end{tabular}

    \caption{
    Qualitative comparison of detector disagreement and post-training fusion across three representative samples.
    Each row shows the input image, ground-truth mask, three candidate detector responses, equal-weight fusion, \method{}, and GT-referenced error maps for the two fusion results.
    The MVTec AD hybrid example exhibits the strongest detector disagreement, while the other two examples show more moderate differences between equal-weight fusion and \method{}.
    }
    \label{fig:qualitative-main}
\end{figure*}
\begin{table*}[t]
\centering
\footnotesize
\setlength{\tabcolsep}{20pt}
\renewcommand{\arraystretch}{0.8}
\caption{Complete comparison on VisA across the three detector pools.
All values are percentages, and the best result within each detector-pool block is boldfaced.}
\label{tab:primary-visa}
\begin{tabular}{lrrrrrr}
\toprule
& \multicolumn{3}{c}{Image-level} & \multicolumn{3}{c}{Pixel-level} \\
\cmidrule(lr){2-4} \cmidrule(lr){5-7}
Method & AUROC & AP & F1 & AUROC & AP & F1 \\
\midrule
\midrule

\multicolumn{7}{c}{\textit{CLIP-based detector pool}} \\
\midrule
AdaCLIP~\cite{cao2024adaclip}
& 93.75 & 94.35 & 89.78
& 95.79 & 28.38 & 38.39 \\
AnomalyCLIP~\cite{zhou2023anomalyclip}
& 81.97 & 85.34 & 80.44
& 95.43 & 14.14 & 23.12 \\
WinCLIP~\cite{jeong2023winclip}
& 81.67 & 83.68 & 81.44
& 88.77 & 15.26 & 22.41 \\
Equal-weight~\cite{dietterich2000ensemble}
& 91.19 & 92.26 & 87.38
& 97.90 & 30.06 & 38.56 \\
Inverse-distance~\cite{shepard1968two}
& 91.11 & 92.35 & 87.22
& 97.68 & 28.98 & 37.95 \\
Voting~\cite{aggarwal2016outlier,rayana2016less,boateng2023unsupervised}
& 88.66 & 90.07 & 85.58
& 93.54 & 18.48 & 31.43 \\
Stacking~\cite{wolpert1992stacked}
& 93.66 & 94.21 & 89.81
& 97.92 & 30.76 & 39.15 \\
\rowcolor{gray!10}
\method{}
& \textbf{93.88} & \textbf{94.43} & \textbf{90.07}
& \textbf{97.95} & \textbf{31.62} & \textbf{40.54} \\

\midrule
\midrule

\multicolumn{7}{c}{\textit{Feature-reconstruction detector pool}} \\
\midrule
MambaAD~\cite{he2024mambaad}
& 94.64 & 95.35 & 90.26
& 98.38 & 26.98 & 37.38 \\
OneNIP~\cite{gao2024learning}
& 93.32 & 95.22 & 90.18
& 98.78 & 28.80 & 38.30 \\
InvAD~\cite{zhang2024invad}
& 95.22 & 96.55 & 91.90
& 97.53 & 20.46 & 30.56 \\
Equal-weight~\cite{dietterich2000ensemble}
& 96.83 & 97.56 & 93.92
& 99.13 & 37.39 & 44.35 \\
Inverse-distance~\cite{shepard1968two}
& 96.78 & 97.45 & 93.54
& \textbf{99.15} & 37.28 & 44.10 \\
Voting~\cite{aggarwal2016outlier,rayana2016less,boateng2023unsupervised}
& 81.65 & 84.66 & 81.04
& 96.91 & 18.50 & 28.57 \\
Stacking~\cite{wolpert1992stacked}
& 96.76 & 97.51 & 93.97
& 98.81 & 29.07 & 40.03 \\
\rowcolor{gray!10}
\method{}
& \textbf{96.92} & \textbf{97.62} & \textbf{94.10}
& \textbf{99.15} & \textbf{37.69} & \textbf{44.86} \\

\midrule
\midrule

\multicolumn{7}{c}{\textit{Hybrid detector pool}} \\
\midrule
OneNIP~\cite{gao2024learning}
& 93.32 & 95.22 & 90.18
& 98.63 & 30.08 & 39.72 \\
InvAD~\cite{zhang2024invad}
& 95.22 & 96.55 & 91.90
& 97.46 & 21.79 & 32.26 \\
AdaCLIP~\cite{cao2024adaclip}
& 93.75 & 94.35 & 89.78
& 95.62 & 30.04 & 40.37 \\
Equal-weight~\cite{dietterich2000ensemble}
& 98.07 & 98.47 & 95.16
& 99.30 & 45.62 & 51.72 \\
Inverse-distance~\cite{shepard1968two}
& 97.73 & 98.23 & 95.02
& 99.32 & 45.36 & 51.76 \\
Voting~\cite{aggarwal2016outlier,rayana2016less,boateng2023unsupervised}
& 92.90 & 94.19 & 90.09
& 97.64 & 29.59 & 41.66 \\
Stacking~\cite{wolpert1992stacked}
& 98.06 & 98.46 & 95.03
& 99.09 & 43.15 & 50.12 \\
\rowcolor{gray!10}
\method{}
& \textbf{98.48} & \textbf{98.83} & \textbf{95.73}
& \textbf{99.33} & \textbf{45.66} & \textbf{51.89} \\

\bottomrule
\end{tabular}
\vspace{-2mm}
\end{table*}
\begin{table*}[!t]
\centering
\footnotesize
\setlength{\tabcolsep}{0.2pt}
\renewcommand{\arraystretch}{0.7}
\caption{Transfer evaluation on seven public anomaly benchmarks.
The compared individual detectors are AdaCLIP~\cite{cao2024adaclip},
AnomalyCLIP~\cite{zhou2023anomalyclip}, and WinCLIP~\cite{jeong2023winclip}.
All values are percentages, and the best results within each benchmark are boldfaced.}
\label{tab:transfer-results}

\begin{tabular}{L{1.8cm}*{8}{C{2.00cm}}}
\toprule
Metric
& AdaCLIP~\cite{cao2024adaclip}
& AnomalyCLIP~\cite{zhou2023anomalyclip}
& WinCLIP~\cite{jeong2023winclip}
& Equal~\cite{dietterich2000ensemble}
& InvDist~\cite{shepard1968two}
& Voting~\cite{aggarwal2016outlier,rayana2016less,boateng2023unsupervised}
& Stacking~\cite{wolpert1992stacked}
& \method{} \\
\midrule
\midrule

\multicolumn{9}{c}{\textit{BTAD}} \\
\midrule
Image AUROC & 91.32 & 89.12 & 75.55 & 94.22 & 93.80 & 91.15 & 94.21 & \textbf{95.31} \\
Image AP    & 94.46 & 91.05 & 68.09 & 95.19 & 94.70 & 93.79 & 95.01 & \textbf{96.08} \\
Image F1    & 88.92 & 85.99 & 67.22 & 89.78 & 90.49 & 88.44 & 90.12 & \textbf{92.96} \\
\addlinespace[1pt]
Pixel AUROC & 93.64 & 93.25 & 95.81 & 97.25 & 96.95 & 92.96 & 96.45 & \textbf{97.30} \\
Pixel AP    & 41.32 & 35.59 & 34.46 & 49.29 & 47.93 & 31.69 & 47.25 & \textbf{49.87} \\
Pixel F1    & 46.56 & 42.84 & 42.07 & 53.09 & 52.35 & 38.75 & 51.33 & \textbf{53.82} \\

\midrule
\midrule

\multicolumn{9}{c}{\textit{KSDD/SDD}} \\
\midrule
Image AUROC & 96.21 & 96.06 & 90.83 & 95.99 & 94.68 & 95.40 & 96.33 & \textbf{96.92} \\
Image AP    & 90.82 & 85.47 & 81.12 & 90.27 & 88.12 & 83.26 & 91.12 & \textbf{91.77} \\
Image F1    & 86.24 & 82.57 & 78.00 & 85.71 & 82.69 & 85.44 & 87.13 & \textbf{89.32} \\
\addlinespace[1pt]
Pixel AUROC & 94.15 & 97.68 & 97.72 & 98.21 & 98.10 & 95.16 & \textbf{98.47} & 98.43 \\
Pixel AP    & 14.90 & 19.79 & 15.93 & 34.17 & 34.02 & 21.28 & 33.49 & \textbf{34.65} \\
Pixel F1    & 29.34 & 34.42 & 29.24 & \textbf{48.97} & \textbf{48.97} & 39.45 & 48.46 & \textbf{48.97} \\

\midrule
\midrule

\multicolumn{9}{c}{\textit{DAGM}} \\
\midrule
Image AUROC & \textbf{99.99} & 97.26 & 96.43 & 99.89 & 99.91 & 99.85 & 99.98 & \textbf{99.99} \\
Image AP    & \textbf{99.94} & 92.19 & 92.95 & 99.61 & 99.70 & 99.45 & \textbf{99.94} & \textbf{99.94} \\
Image F1    & 99.83 & 91.68 & 89.99 & 98.79 & 99.08 & 98.74 & 99.83 & \textbf{99.89} \\
\addlinespace[1pt]
Pixel AUROC & \textbf{79.95} & 59.79 & 63.35 & 68.70 & 72.33 & 44.56 & 70.66 & 74.03 \\
Pixel AP    & \textbf{0.95} & 0.44 & 0.48 & 0.65 & 0.68 & 0.38 & 0.67 & 0.74 \\
Pixel F1    & 2.72 & 2.09 & 1.66 & 2.85 & 2.85 & 2.80 & 2.73 & \textbf{2.86} \\

\midrule
\midrule

\multicolumn{9}{c}{\textit{DTD}} \\
\midrule
Image AUROC & 99.16 & 94.46 & 97.26 & 99.26 & 99.09 & 98.63 & 99.26 & \textbf{99.41} \\
Image AP    & 99.66 & 97.70 & 98.80 & 99.67 & 99.56 & 99.16 & 99.70 & \textbf{99.74} \\
Image F1    & 98.96 & 94.53 & 96.18 & 98.58 & 98.36 & 98.39 & 99.08 & \textbf{99.25} \\
\addlinespace[1pt]
Pixel AUROC & \textbf{99.48} & 97.52 & 98.20 & 99.11 & 99.11 & 98.08 & 99.10 & 99.20 \\
Pixel AP    & 72.71 & 44.49 & 51.61 & 73.00 & 72.41 & 61.96 & 74.36 & \textbf{75.12} \\
Pixel F1    & 76.80 & 54.94 & 60.69 & 74.47 & 73.42 & 71.20 & 76.02 & \textbf{76.82} \\

\midrule
\midrule

\multicolumn{9}{c}{\textit{Magnetic Tile}} \\
\midrule
Image AUROC & 74.62 & 70.26 & 67.35 & 73.87 & 72.74 & 72.96 & 75.17 & \textbf{77.23} \\
Image AP    & 60.88 & 54.05 & 50.61 & 59.42 & 58.10 & 52.28 & 61.02 & \textbf{62.22} \\
Image F1    & 55.36 & 51.94 & 49.68 & 53.71 & 52.83 & 53.91 & 55.65 & \textbf{57.57} \\
\addlinespace[1pt]
Pixel AUROC & 75.49 & 69.50 & 72.87 & 75.97 & 75.60 & 67.77 & 75.66 & \textbf{76.74} \\
Pixel AP    & \textbf{17.07} & 6.80 & 15.49 & 14.26 & 13.00 & 6.69 & 14.42 & 15.61 \\
Pixel F1    & 24.10 & 12.86 & 19.21 & 21.53 & 19.81 & 12.65 & 22.34 & \textbf{24.42} \\

\midrule
\midrule

\multicolumn{9}{c}{\textit{KolektorSDD2}} \\
\midrule
Image AUROC & 95.85 & 90.81 & 93.50 & 95.61 & 95.47 & 95.43 & 95.51 & \textbf{96.49} \\
Image AP    & 88.99 & 76.46 & 75.27 & 88.98 & 88.26 & 88.55 & 89.21 & \textbf{91.15} \\
Image F1    & 85.17 & 71.79 & 71.37 & 85.02 & 82.18 & 85.02 & 85.57 & \textbf{87.44} \\
\addlinespace[1pt]
Pixel AUROC & 98.42 & 97.87 & 84.20 & 98.87 & \textbf{99.03} & 94.25 & 98.91 & 98.97 \\
Pixel AP    & \textbf{69.19} & 47.70 & 5.18 & 45.81 & 45.95 & 31.12 & 43.65 & 46.66 \\
Pixel F1    & \textbf{66.65} & 54.13 & 13.04 & 57.22 & 57.09 & 52.83 & 57.08 & 57.75 \\

\midrule
\midrule

\multicolumn{9}{c}{\textit{MVTec LOCO}} \\
\midrule
Image AUROC & 59.59 & 61.02 & 59.36 & 63.98 & 62.79 & 62.69 & 61.15 & \textbf{64.01} \\
Image AP    & 72.81 & 74.66 & 71.16 & 75.85 & 75.29 & 72.86 & 73.59 & \textbf{75.95} \\
Image F1    & 76.95 & 76.87 & 76.96 & 77.15 & 77.07 & 76.63 & 76.92 & \textbf{77.22} \\
\addlinespace[1pt]
Pixel AUROC & 58.86 & 64.21 & \textbf{72.81} & 72.35 & 71.29 & 69.86 & 72.61 & \textbf{72.81} \\
Pixel AP    & 9.20 & 10.22 & \textbf{15.48} & 12.90 & 12.37 & 12.37 & 13.11 & \textbf{15.48} \\
Pixel F1    & 16.05 & 16.85 & 21.41 & 21.07 & 20.15 & \textbf{22.83} & 21.29 & 21.41 \\

\midrule
\midrule

\multicolumn{9}{c}{\textit{Average over seven benchmarks}} \\
\midrule
Image AUROC & 88.10 & 85.57 & 82.90 & 88.97 & 88.35 & 88.02 & 88.80 & \textbf{89.91} \\
Image AP    & 86.80 & 81.66 & 76.86 & 87.00 & 86.25 & 84.19 & 87.08 & \textbf{88.12} \\
Image F1    & 84.49 & 79.34 & 75.63 & 84.11 & 83.24 & 83.80 & 84.90 & \textbf{86.24} \\
\addlinespace[1pt]
Pixel AUROC & 85.71 & 82.83 & 83.57 & 87.21 & 87.49 & 80.38 & 87.41 & \textbf{88.21} \\
Pixel AP    & 32.19 & 23.58 & 19.81 & 32.87 & 32.33 & 23.64 & 32.42 & \textbf{34.02} \\
Pixel F1    & 37.46 & 31.16 & 26.76 & 39.89 & 39.23 & 34.36 & 39.89 & \textbf{40.86} \\

\bottomrule
\end{tabular}
\vspace{-2mm}
\end{table*}
\subsection{Constrained VLM Fusion Policy}

We formulate the VLM as an agent that observes a structured evidence state and selects score- and map-level actions from a bounded fusion-action space.
For each selected sample $i\in\mathcal{I}_{\tau}$, the VLM receives a structured visual input together with a textual evidence summary derived from $E_i$. The visual input contains the original image, individual detector maps, the equal-weight reference, detector disagreement, and candidate-region overlays under a common normalization. The textual summary provides detector scores, map statistics, conflict cues, candidate regions, and the valid fusion actions.

Rather than generating a new anomaly score or mask, the VLM acts as a constrained selector over two finite action spaces:
$\mathcal{A}_i^s$ for image-score fusion and $\mathcal{A}_i^m$ for pixel-map fusion. Its policy is written as
\begin{equation}
    (a_i^s,a_i^m)=
    \pi_{\mathrm{VLM}}
    \left(E_i,\mathcal{A}_i^s,\mathcal{A}_i^m\right),
\end{equation}
where $a_i^s\in\mathcal{A}_i^s$ and $a_i^m\in\mathcal{A}_i^m$ denote the selected score- and map-level actions, respectively.

The available action families are summarized in Table~\ref{tab:action-space}. Image-level actions include score consensus, robust aggregation, detector weighting, and score--map consistency, while pixel-level actions include map consensus, robust aggregation, detector weighting, geometric consensus, and disagreement correction. In the score--map consistency family, $M$ denotes the map summary paired with the
score action. The weights $w_k$ are nonnegative and sum to one, while $\alpha$, $\eta$, and $\epsilon$ control score--map coupling, disagreement correction, and numerical stability, respectively.

The VLM returns only valid action identifiers together with an optional short rationale used for interpretation. The parser accepts only identifiers from the predefined action sets; invalid or incomplete outputs revert to the fallback action. Consequently, the VLM can use the original image and detector evidence to determine how the predictions should be fused, but the final score and anomaly map always remain deterministic transformations of the pretrained detector outputs.

\subsection{Final Prediction}

The final prediction combines the fallback fusion with VLM-selected actions:
\begin{align}
\hat{s}_i &=
\begin{cases}
F_s(a_i^s;\tilde{s}_{i1:K},\tilde{M}_{i1:K}), & i\in\mathcal{I}_{\tau},\\
s_i^0, & i\notin\mathcal{I}_{\tau},
\end{cases}
\nonumber\\
\hat{M}_i &=
\begin{cases}
F_m(a_i^m;\tilde{M}_{i1:K}), & i\in\mathcal{I}_{\tau},\\
M_i^0, & i\notin\mathcal{I}_{\tau}.
\end{cases}
\end{align}

All components of the fusion protocol, including the fallback actions, VLM prompt, parser, action space, and conflict-triggering rule, are fixed before evaluation. Therefore, \method{} requires neither anomaly labels nor detector retraining or task-specific meta-training at evaluation time.
\section{Experiments}
\label{sec:experiments}

\subsection{Experimental Settings}

\paragraph{Benchmarks and metrics.}
We evaluate \method{} on MVTec AD~\citep{bergmann2019mvtec}, VisA~\citep{zou2022spot}, and seven additional benchmarks: BTAD~\cite{mishra2021vt}, KSDD/SDD~\cite{tabernik2020segmentation}, DAGM~\cite{wieler2007dagm}, DTD~\cite{cimpoi2014describing}, Magnetic Tile~\cite{huang2020surface}, KolektorSDD2~\cite{bovzivc2021mixed}, and MVTec LOCO~\cite{bergmann2022beyond}. Real-IAD~\citep{wang2024real} is additionally used as a large-scale diagnostic for the CLIP-based detector pool.
We report macro-averaged image AUROC, image AP, image F1, pixel AUROC, pixel AP, and pixel F1, all in percentage.

\paragraph{Detector pools.}
On MVTec AD and VisA, we evaluate three heterogeneous three-detector pools. The CLIP-based pool contains AdaCLIP~\citep{cao2024adaclip}, AnomalyCLIP~\citep{zhou2023anomalyclip}, and WinCLIP~\citep{jeong2023winclip}. 
The feature-reconstruction pool contains MambaAD~\citep{he2024mambaad}, OneNIP~\citep{gao2024learning}, and InvAD~\citep{zhang2024invad}. 
The hybrid pool contains OneNIP, InvAD, and AdaCLIP. 
These pools cover CLIP-based anomaly detectors, feature- or reconstruction-based detectors, and cross-paradigm detector combinations.

\paragraph{Baselines.}
The comparison includes the three individual candidate detectors in each pool, equal-weight fusion~\cite{dietterich2000ensemble}, inverse-distance fusion, voting-based fusion~\cite{aggarwal2016outlier,rayana2016less,boateng2023unsupervised}, and stacking-style fusion~\cite{wolpert1992stacked}. 
Equal-weight fusion uses the same sample alignment, map resolution, and percentile normalization as \method{}, and then averages image scores and pixel maps. 
The other baselines provide training-free alternatives over the same saved detector outputs. 
The stacking-style baseline is an unsupervised score-statistic heuristic rather than a supervised meta-learner. 
Box fusion~\cite{solovyev2021weighted} and Vote-NMS~\cite{neubeck2006efficient} are reported separately as diagnostic map-to-box adaptations because they discard dense localization structure.

\paragraph{Implementation details.}
For each benchmark setting, the conflict score in Eq.~\eqref{eq:conflict} is computed for all samples, and we use a fixed intervention threshold of $\tau=0.5$.
Samples with conflict scores above the threshold select one image-score action and one pixel-map action from the predefined action bank, while the remaining samples use a fixed setting-specific fallback action.
Invalid VLM outputs revert to the corresponding fallback action.
We use LLaVA-7B as the VLM backend with temperature $0.1$ and a maximum generation length of 160 tokens.
All detector outputs are normalized by 1st--99th percentile scaling within category using the unlabeled evaluation split, and the same normalization is applied to all ensemble baselines.
Dense anomaly maps are resized to the ground-truth mask resolution before pixel-level evaluation.
The conflict score uses $\lambda_s=\lambda_m=0.35$ and $\lambda_p=\lambda_r=0.15$, with a top-response fraction of $q=0.01$.
The finite action bank uses predefined detector-weighting vectors, score--map coupling coefficients $\alpha\in\{0.15,0.20,0.40\}$, disagreement-correction strengths $\eta\in\{0.02,0.15\}$, and $\epsilon=10^{-6}$ for geometric aggregation.
No anomaly labels are used for evaluation-time normalization, triggering, or per-sample VLM action selection.

\paragraph{Constrained agent prompt.}
\method{} uses a constrained prompt interface.
The VLM receives detector evidence and selects one image-level and one pixel-level fusion action from the predefined action sets.
The prompt is instantiated with the dataset, category, detector scores, conflict statistics, candidate tools, and valid fusion actions.

\section{Results}
\label{sec:results}

\subsection{Primary Benchmark Results}

Tables~\ref{tab:primary-mvtec} and~\ref{tab:primary-visa} report six-metric comparisons on MVTec AD and VisA across three detector pools.
All ensemble methods use the same saved detector outputs, category-wise normalization, and map resolution.
The strongest individual detector varies across datasets, pools, and metrics, while fixed fusion rules do not consistently improve both image- and pixel-level performance.

\method{} improves over equal-weight fusion in all six primary configurations, with the largest gain on MVTec AD using the CLIP-based pool, especially for pixel-level localization.
These results show that detector reliability varies across both images and spatial responses, motivating adaptive score- and map-level fusion.

\subsection{Transfer and Large-Scale Evaluation}

We further evaluate on BTAD~\cite{mishra2021vt}, KSDD/SDD~\cite{tabernik2020segmentation}, DAGM~\cite{wieler2007dagm}, DTD~\cite{cimpoi2014describing}, Magnetic Tile~\cite{huang2020surface}, KolektorSDD2~\cite{bovzivc2021mixed}, and MVTec LOCO~\cite{bergmann2022beyond}.
As shown in Table~\ref{tab:transfer-results}, \method{} consistently improves over equal-weight fusion and remains competitive with or stronger than other fixed fusion strategies without detector retraining or dataset-specific fitting.

\paragraph{Large-scale Real-IAD evaluation.}
On the full Real-IAD benchmark~\cite{wang2024real}, Table~\ref{tab:realiad-results} shows that \method{} improves image AUROC/AP/F1 from 72.47/58.85/59.66 to 75.53/62.54/62.49, confirming its effectiveness at larger scale.

\begin{table}[!t]
\centering
\footnotesize
\setlength{\tabcolsep}{14pt}
\renewcommand{\arraystretch}{0.9}
\caption{Large-scale score-level evaluation on Real-IAD with the CLIP-based detector pool.
All values are percentages, and the best result is boldfaced.}
\label{tab:realiad-results}
\begin{tabular}{lrrr}
\toprule
Method & AUROC & AP & F1 \\
\midrule
AdaCLIP~\cite{cao2024adaclip}
& 74.15 & 61.44 & 60.88 \\
AnomalyCLIP~\cite{zhou2023anomalyclip}
& 68.60 & 53.28 & 56.32 \\
WinCLIP~\cite{jeong2023winclip}
& 63.45 & 45.88 & 57.25 \\
Equal-weight~\cite{dietterich2000ensemble}
& 72.47 & 58.85 & 59.66 \\
\method{}
& \textbf{75.53} & \textbf{62.54} & \textbf{62.49} \\
\bottomrule
\end{tabular}
\end{table}

\subsection{Robustness and Diagnostic Analysis}

\paragraph{Detector complementarity.}
Table~\ref{tab:complementarity} reports pairwise score correlations and high-response-region overlap for the CLIP-based pool.

\begin{table}[t]
\centering
\footnotesize
\renewcommand{\arraystretch}{0.9}
\caption{Pairwise complementarity of AdaCLIP, AnomalyCLIP, and WinCLIP on MVTec AD and VisA.
Score correlation is measured by Spearman correlation, while spatial overlap is measured by the IoU of high-response regions.
Lower IoU indicates more diverse spatial responses.}
\label{tab:complementarity}
\setlength{\tabcolsep}{2pt}
\begin{tabular}{llcc}
\toprule
Dataset & Detector Pair & Score Spearman & Top-response IoU \\
\midrule
\multirow{3}{*}{MVTec AD}
& AdaCLIP--AnomalyCLIP & 0.615 & 0.160 \\
& AdaCLIP--WinCLIP     & 0.502 & 0.163 \\
& AnomalyCLIP--WinCLIP & 0.511 & 0.114 \\
\midrule
\multirow{3}{*}{VisA}
& AdaCLIP--AnomalyCLIP & 0.761 & 0.182 \\
& AdaCLIP--WinCLIP     & 0.472 & 0.048 \\
& AnomalyCLIP--WinCLIP & 0.436 & 0.042 \\
\bottomrule
\end{tabular}
\end{table}

The detectors show moderate score agreement but substantially different spatial responses.
On VisA, WinCLIP has top-response IoUs of only 0.048 and 0.042 with AdaCLIP and AnomalyCLIP, respectively, supporting the need for adaptive fusion.

\paragraph{Calibration robustness.}
Table~\ref{tab:calibration} compares default transductive calibration with 1st--99th percentile statistics estimated from 64 stored samples per category.
Performance remains comparable or improves, indicating that \method{} is not critically dependent on the full evaluation split.

\begin{table}[t]
\centering
\footnotesize
\renewcommand{\arraystretch}{0.9}
\caption{Calibration robustness for the CLIP-based detector pool.
Fixed-prefix calibration estimates the 1st--99th percentile bounds from 64 stored samples per category without using anomaly labels.}
\label{tab:calibration}
\setlength{\tabcolsep}{6pt}
\begin{tabular}{lccc}
\toprule
Dataset & Calibration & Image AUROC & Pixel AP \\
\midrule
\multirow{2}{*}{MVTec AD} & Transductive & 98.84 & 64.42 \\
         & Fixed prefix (64) & 98.75 & 64.71 \\
\midrule
\multirow{2}{*}{VisA}     & Transductive & 88.19 & 30.14 \\
         & Fixed prefix (64) & 89.16 & 32.86 \\
\bottomrule
\end{tabular}
\end{table}

\paragraph{Calibration-size sensitivity.}
We further vary the number of unlabeled calibration samples on MVTec AD.

\begin{table}[t]
\centering
\footnotesize
\caption{Calibration-size sensitivity of \method{} on the MVTec AD CLIP-based detector pool.
Normalization statistics are estimated from a fixed unlabeled prefix of each category and then kept frozen.}
\label{tab:calibration-size}
\setlength{\tabcolsep}{4pt}
\begin{tabular}{c|cccccc}
\toprule
Size & Im. AUROC & Im. AP & Im. F1 & Pix. AUROC & Pix. AP & Pix. F1 \\
\midrule
16  & 97.77 & 98.97 & 96.81 & 97.98 & 63.90 & 65.44 \\
32  & 98.28 & 99.22 & 97.20 & 97.92 & 64.24 & 65.26 \\
64  & 98.75 & 99.48 & 97.80 & 98.05 & 64.71 & 65.62 \\
128 & 98.80 & 99.49 & 97.90 & 97.84 & 64.24 & 65.27 \\
\bottomrule
\end{tabular}
\end{table}

Performance improves from 16 to 32 samples and remains stable for 32--128 samples, showing that reliable normalization requires only limited calibration data.

\paragraph{Action-bank diagnostic.}
We progressively enlarge the fusion-action bank on the same replayed triggered samples.

\begin{table}[!t]
\centering
\footnotesize
\caption{Diagnostic action-bank replay on MVTec AD with the CLIP-based detector pool.
All variants use the same replayed triggered samples; unavailable historical actions are deterministically remapped to valid actions in the restricted bank.}
\label{tab:action-bank}
\setlength{\tabcolsep}{2pt}
\begin{tabular}{lccc}
\toprule
Action Bank & Pixel AP & Pixel F1 & Newly Used Action \\
\midrule
A1: Equal + Median              & 64.433 & 64.867 & -- \\
A2: + Detector Weighting        & 64.471 & 64.905 & Weighting \\
A3: + Geometric/Consistency   & 64.471 & 64.905 & None \\
A4: Full Bank                   & 64.485 & 64.918 & Disagreement correction \\
\bottomrule
\end{tabular}
\end{table}

Detector weighting yields a small gain over consensus-only actions, while the full bank provides a further modest improvement; geometric consensus and score--map consistency are not selected in this replay.

\paragraph{Selector diagnostic.}
Table~\ref{tab:selector-ablation-visa-clip} compares fallback-only, random, deterministic-rule, and historical VLM selection on VisA.

\begin{table*}[t]
\centering
\footnotesize
\setlength{\tabcolsep}{8pt}
\renewcommand{\arraystretch}{0.9}
\caption{Diagnostic selector replay on VisA with the CLIP-based detector pool.
Fallback-only, random, and deterministic-rule variants use the same reconstructed conflict-trigger set and intervention budget.
Random selection reports mean $\pm$ standard deviation over eight fixed seeds.
}
\label{tab:selector-ablation-visa-clip}
\begin{tabular}{lllllll}
\toprule
Selector
& Image AUROC & Image AP & Image F1
& Pixel AUROC & Pixel AP & Pixel F1 \\
\midrule
Fallback only
& 91.26 & 92.41 & 87.34
& 97.84 & 30.22 & 38.45 \\

Random action selector
& 91.24 $\pm$ 0.04
& 92.37 $\pm$ 0.02
& 87.21 $\pm$ 0.05
& 97.75 $\pm$ 0.14
& 30.09 $\pm$ 0.15
& 38.34 $\pm$ 0.12 \\

Deterministic rule selector
& 92.72 & 93.01 & 89.84
& 97.90 & 30.91 & 39.21 \\

\midrule
Historical VLM result
& \textbf{93.88} & \textbf{94.43} & \textbf{90.07}
& \textbf{97.95} & \textbf{31.62} & \textbf{40.54} \\
\bottomrule
\end{tabular}
\end{table*}

The deterministic selector improves over fallback and random selection, while the historical VLM result performs best.
Because reconstructed trigger indices do not exactly match the original VLM trace, this is a diagnostic replay rather than a strict attribution of gains to VLM reasoning.

\paragraph{Setting-level robustness.}
Table~\ref{tab:setting_bootstrap_ci} reports descriptive resampling over the six primary benchmark--pool settings; the ranges characterize setting-level rather than image-level variation.

\begin{table*}[t]
\centering
\scriptsize
\caption{Setting-level resampling summary on the six primary MVTec AD / VisA detector-pool settings. Values are percentage-point deltas; brackets show 95\% percentile ranges over 10,000 setting-level resamples.}
\label{tab:setting_bootstrap_ci}
\resizebox{\linewidth}{!}{%
\begin{tabular}{lccccc}
\toprule
Comparison & Image AUROC & Image AP & Pixel AUROC & Pixel AP & Pixel F1 \\
\midrule
Direct VLM $-$ Equal & +0.627 [+0.135, +1.489] & +0.447 [+0.033, +1.160] & +0.244 [+0.040, +0.592] & +1.981 [+0.548, +4.175] & +1.793 [+0.607, +3.568] \\
\bottomrule
\end{tabular}%
}
\end{table*}

\paragraph{Diagnostic box-based ensembles.}
We additionally convert dense anomaly maps into boxes and evaluate WBF and Vote-NMS.

\begin{table}[t]
\centering
\small
\caption{Diagnostic box-based fusion baselines averaged over the six MVTec AD and VisA settings.
Values denote the performance difference relative to equal-weight fusion in percentage points.}
\label{tab:box-baselines}
\resizebox{\linewidth}{!}{%
\begin{tabular}{lrrrrrr}
\toprule
Baseline & Image AUROC & Image AP & Image F1 & Pixel AUROC & Pixel AP & Pixel F1\\
\midrule
WBF & -6.556 & -5.488 & -4.861 & -2.789 & -23.498 & -17.579\\
Vote-NMS & -16.447 & -14.258 & -8.519 & -3.817 & -30.400 & -24.512\\
\bottomrule
\end{tabular}%
}
\end{table}

Both substantially underperform equal-weight fusion, particularly on pixel-level metrics, indicating that box-based fusion is poorly suited to dense anomaly localization.

\paragraph{VLM invocation cost.}
Because all methods reuse saved detector predictions, we report only the additional VLM calls introduced by \method{}.
Only high-conflict samples invoke the VLM, while other samples use deterministic fusion and failed calls revert to valid fallback actions.

\begin{table}[!t]
\centering
\footnotesize
\caption{Fusion-stage compute accounting. Detector inference is excluded because all fusion methods replay the same saved detector dumps. The VLM backend is LLaVA-7B.}
\label{tab:compute_accounting}
\setlength{\tabcolsep}{1pt}
\renewcommand{\arraystretch}{0.9}

\begin{tabular}{llr}
\toprule
Experiment & Scope & VLM calls \\
\midrule
Main Direct VLM-Agent
& MVTec AD / VisA primary settings
& 489 \\
Transfer Direct VLM-Agent
& 7 additional settings
& 1286 \\
\bottomrule
\end{tabular}
\end{table}

\subsection{Qualitative Analysis}

Figure~\ref{fig:qualitative-main} shows three representative cases.
When detector predictions agree, \method{} makes only limited changes to equal-weight fusion; under stronger disagreement, it suppresses unsupported diffuse responses while preserving compact anomaly evidence.
These examples illustrate how adaptive fusion resolves detector disagreement without generating new anomaly maps.

\subsection{Discussion and Limitations}
\label{sec:limitations}

Across detector pools and benchmarks, \method{} consistently improves over fixed label-free fusion strategies, with particularly strong gains in localization metrics and robust transfer to additional anomaly benchmarks.
The method is also insensitive to reduced calibration data and keeps each fusion decision traceable to detector outputs, conflict statistics, and selected actions.
Its main limitation is the dependence on the quality and diversity of the candidate detectors; future work may explore broader action spaces and additional VLM backends.
\section{Conclusion}
\label{sec:conclusion}

We presented \method{}, a post-training vision-language agent for anomaly-detector fusion.
Rather than training an additional detector, \method{} normalizes the outputs of pretrained detectors, identifies conflicting evidence, and uses a constrained VLM to select among predefined image-score and pixel-map fusion actions.
Across MVTec AD, VisA, and seven additional public anomaly benchmarks, the method provides consistent improvements over fixed label-free fusion strategies while preserving the original detector outputs and avoiding detector retraining.
These results suggest that post-training anomaly ensembles can benefit not only from stronger base detectors, but also from explicitly resolving heterogeneous confidence scales, spatial responses, and detector disagreement.

\bibliographystyle{IEEEtran}
\bibliography{references}
\end{document}